\documentclass[preprint,12pt]{elsarticle}

\usepackage{amssymb}
\usepackage{amsmath} 
\usepackage{color}
\usepackage{graphicx}
\usepackage{subfig}
\usepackage{multirow}
\usepackage{centernot}
\usepackage{setspace}

\journal{Pattern Recognition}

\begin{document}

\newcommand{\cpos}{\text{\scriptsize{{\it POS}}}}
\newcommand{\cneg}{\text{\scriptsize{{\it NEG}}}}
\newcommand{\mb}{\mathbf}

\doublespacing

\begin{frontmatter}

\title{A bag-to-class divergence approach to multiple-instance learning}

\author[add1]{Kajsa M{\o}llersen}
\ead{kajsa.mollersen@uit.no}
\author[add2]{Jon Yngve Hardeberg}
\author[add3]{Fred Godtliebsen}

\address[add1]{Department of Community Medicine, UiT The Arctic University of Norway, Troms{\o}, Norway}
\address[add2]{Department of Computer Science, Faculty of Information Technology and Electrical Engineering,
NTNU, Norway}
\address[add3]{Department of Mathematics and Statistics, UiT The Arctic University of Norway, Troms{\o}, Norway}

\begin{abstract} 
In multi-instance (MI) learning, each object (bag) consists of multiple feature vectors (instances), and is most commonly regarded as a set of points in a multidimensional space. 
A different viewpoint is that the instances are realisations of random vectors with corresponding probability distribution, and that a bag is the distribution, not the realisations. 
In MI classification, each bag in the training set has a class label, but the instances are unlabelled. 
By introducing the probability distribution space to bag-level classification problems, dissimilarities between probability distributions (divergences) can be applied.
The bag-to-bag Kullback-Leibler information is asymptotically the best classifier, but the typical sparseness of MI training sets is an obstacle. 
We introduce bag-to-class divergence to MI learning, emphasising the hierarchical nature of the random vectors that makes bags from the same class different. 
We propose two properties for bag-to-class divergences, and an additional property for sparse training sets. 
\end{abstract}


\end{frontmatter}

\clearpage

\section{Introduction} \label{sec:Introduction} 

\subsection{Multi-instance learning} 

In supervised learning, the training data consists of $K$ objects, $\mb{x}$, with corresponding class labels, $y$; $\{(\mb{x}_1,y_1), \ldots, (\mb{x}_k,y_k), \ldots, (\mb{x}_K,y_K)\}$. 
An object is typically a vector of $d$ feature values, $\mb{x}_k = (x_{k1}, \ldots, x_{kd})$, named {\it instance}. 
In multi-instance (MI) learning, each object consists of several instances. 
The set $\mathbb{X}_k = \{ \mb{x}_{k1}, \ldots, \mb{x}_{kn_{k}} \}$, where the $n_k$ elements are vectors of length $d$, is referred to as {\it bag}. 
The number of instances, $n_k$, varies from bag to bag, whereas the vector length is constant. 
In supervised MI learning, the training data consists of $K$ sets and their corresponding class labels,  $\{(\mathbb{X}_1, y_1), \ldots, (\mathbb{X}_k, y_k), \ldots, (\mathbb{X}_K,y_K)\}$. 

\begin{figure}[t!]
    \centering
    \subfloat[Benign]{
      \includegraphics[width=0.3\textwidth]{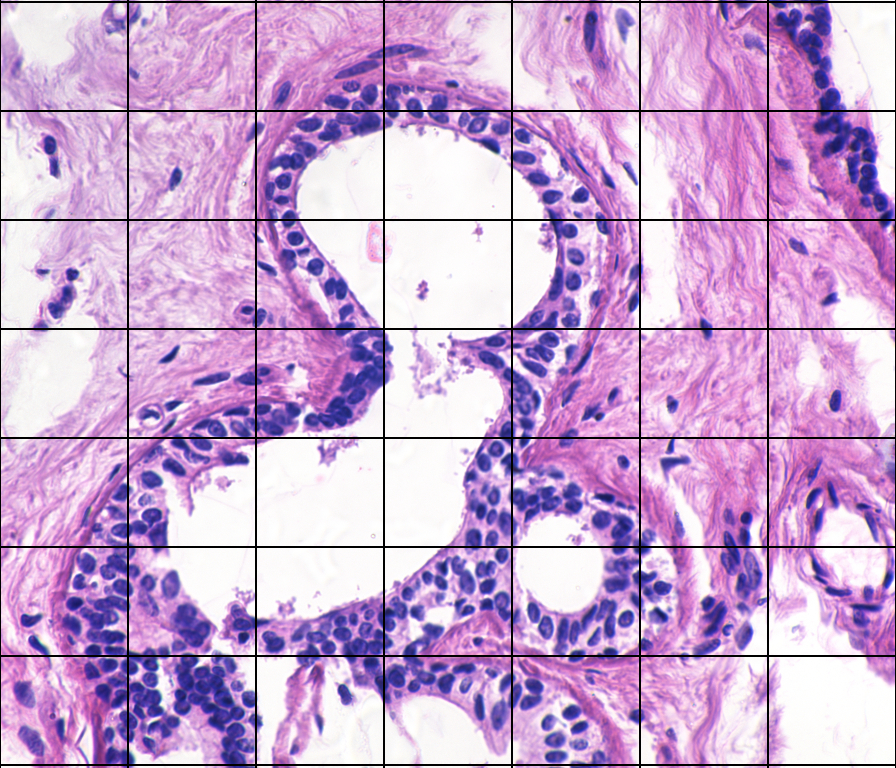} \label{fig:Benign}}
    \subfloat[Malignant]{
        \includegraphics[width=0.3\textwidth]{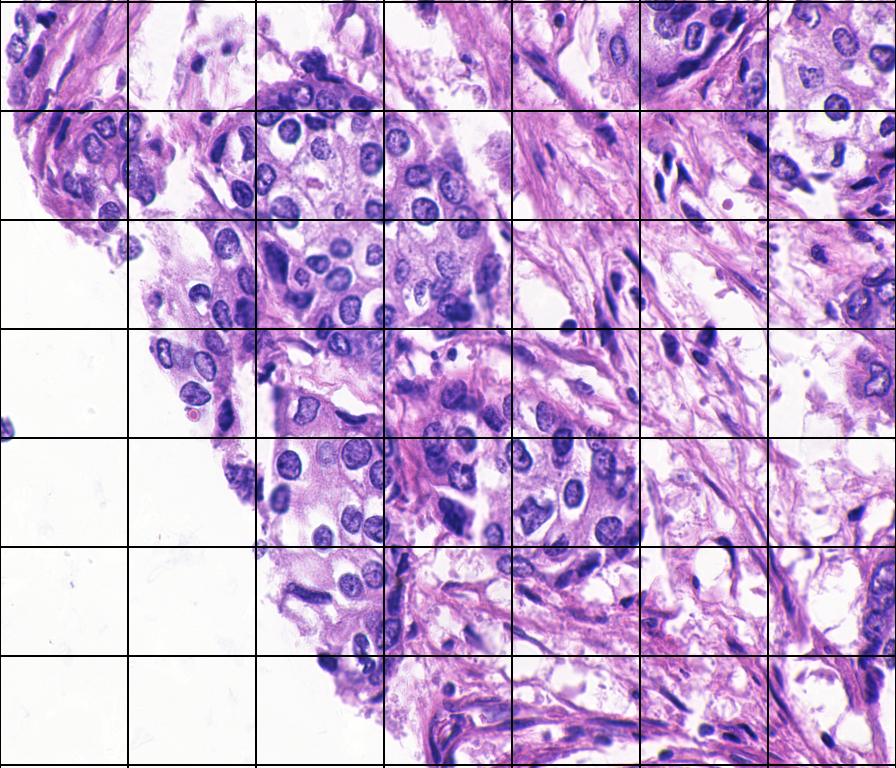} \label{fig:Malignant}}
        \caption{Breast tissue images. The image segments are not labelled.}
    \label{fig:Breast}
\end{figure}

Figure~\ref{fig:Benign} shows an image (bag), $k$, of benign breast tissue \cite{Gelasca2008Evaluation}, divided into $n_k$ segments with corresponding feature vectors (instances) $\mb{x}_{k1}, \ldots, \mb{x}_{kn_k}$ \cite{Kandemir2014Empowering}. 
Correspondingly, figure~\ref{fig:Malignant} shows malignant breast tissue. 
The images in the data set have class labels, the individual segments do not.
This is a key characteristic of MI learning: the instances are not labelled. 
MI learning includes instance classification \cite{Doran2016MultipleInstance}, clustering \cite{Zhang2009Multiinstance}, regression \cite{Zhang2009Multiinstance}, and multi-label learning \cite{Zhou2012Multiinstance, Tang2017Deep}, but this article will focus on bag classification. 
MI learning can also be found as integrated parts of end-to-end methods for image analysis that generate patches, extract features and do feature selection \cite{Tang2017Deep}. 
See also \cite{Wang2018Revisiting} for an overview and discussion on end-to-end neural network MI learning methods. 

The term MI learning was introduced in an application of molecules (bags) with different shapes (instances), and their ability to bind to other molecules \cite{Dietterich1997Solving}. 
A molecule binds if at least one of its shapes can bind.
In MI terminology, the classes, $C$, in binary classification are referred to as positive, $pos$, and negative, $neg$. 
The assumption that a positive bag contains at least one positive instance, and a negative bag contains only negative instances is referred to as the standard MI assumption. 

Many new applications violate the standard MI assumption, such as image classification \cite{Xu2016Multipleinstance} and text categorisation \cite{Qiao2017Diversified}. 
Consequently, successful algorithms meet more general assumptions, see e.g.\ the hierarchy of Weidmann et al.~\cite{Weidmann2003Twolevel} or Foulds and Frank's taxonomy \cite{Foulds2010Review}. 
For a more recent review on MI classification algorithms, see e.g.\ \cite{Cheplygina2015Multiple}.
Carbonneau et al.~\cite{Carbonneau2018Multiple} discussed sample independence and data sparsity, which we address in Section~\ref{sec:Bagtoclass}.
Amores \cite{Amores2013Multiple} presented the three paradigms of instance space (IS), embedded space (ES), and bag space (BS). 
IS methods aggregate the outcome of single-instance classifiers applied to the instances of a bag, whereas ES methods map the instances to a vector, and then use a single-instance classifier.
In the BS paradigm, the instances are transformed to a non-vectorial space where the classification is performed, avoiding the detour via single-instance classifiers.
The non-vectorial space of probability functions has not yet been introduced to the BS paradigm, despite its analytical benefits. 

Whereas both Carbonneau et al.~\cite{Carbonneau2018Multiple} and Amores \cite{Amores2013Multiple} defined a bag as a set of feature vectors, Foulds and Frank \cite{Foulds2010Review} stated that a bag can also be modelled as a probability distribution. 
The distinction is necessary in analysis of classification approaches, and both viewpoints offer benefits, see Section~\ref{sec:Point} for a discussion.

\subsection{The non-vectorial space of probability functions} \label{sec:ProbSpace}

From the probabilistic viewpoint, an instance is a realisation of a random vector, $X$, with probability distribution $P(X)$ and sample space $\mathcal{X}$.
The posterior probability, $P(C|\mathbb{X}_k)$, is an effective classifier if the standard MI assumption holds, since it is known beforehand to be 
\begin{align*} 
  \begin{split}
  P(pos|\mathbb{X}_k) &= \begin{cases} 1 \text{ if any } \mb{x}_{ki} \in \mathcal{X}^+, \, i = 1, \ldots, n_k\\
                                           0 \text{ otherwise, } \end{cases}
  \end{split}                                
\end{align*}            
where $\mathcal{X}^+$ is the positive instance space, and the positive and negative instance spaces are disjoint.               

Bayes' rule, $P(C|X) \propto P(X|C)P(C)$, can be used when the posterior probability is unknown. 
An assumption used to estimate the probability distribution of instance given the class, $P(X|C)$, is that instances from bags of the same class are independent and identically distributed (i.i.d.) random samples, but this is a poor description for MI learning.
As an illustrative example, let the instances be the colour of image segments from the class {\it sea}. 
Image $k$ depicts a clear blue sea, whereas image $\ell$ depicts a deep green sea, and instance distributions are clearly dependent not only on class, but also on bag. 
The random vectors in $\mathbb{X}_k$ are i.i.d., but have a different distribution than those in $\mathbb{X}_{\ell}$.
An important distinction between uncertain objects, whose distribution depends solely on the class label \cite{Jiang2013Clustering, Kriegel2005Densitybased}, and MI learning is that the instances of two bags from the same class are not from the same distribution. 

The dependency nature for MI learning can be described as a hierarchical distribution (Eq.~\ref{eq:GeneralHierarchical}), where a bag, $B$, is defined as the probability distribution of its instances, $P(X|B)$, and the bag space, $\mathcal{B}$, is a set of distributions.

\subsection{Dissimilarities in MI learning} 

Dissimilarities in MI learning can be categorised as instance-to-instance, bag-to-bag or bag-to-class.
Amores \cite{Amores2013Multiple} implicitly assumed metricity for dissimilarity functions \cite{Scholkopf2000Kernel} in the BS paradigm, but there is nothing inherent to MI learning that imposes these restrictions. 
The non-metric Kullback-Leibler (KL) information \cite{Kullback1951Information} is an example of a divergence: a dissimilarity measure between two probability distributions.

Divergences have not been used in MI learning, due to the lack of a probability function space defined for the BS paradigm, despite the benefit of analysis independent of specific data sets \cite{Gibbs2002Choosing}.  
The $f$-divergences \cite{Ali1966General,Csiszar1967Informationtype} have desirable properties for dissimilarity measures, including minimum value for equal distributions, but there is no complete categorisation of divergences. 
The KL information is a non-symmetric $f$-divergence, often used in both statistics and computer science, and is defined as follows for two probability density functions (pdfs) $f_k(\mb{x})$ and $f_\ell(\mb{x})$:
\begin{align} \label{eq:KLinformation}
  D_{KL}(f_k, f_\ell) =  \int f_k(\mb{x}) \log \frac{f_k(\mb{x})}{f_\ell(\mb{x})} d\mb{x}.
\end{align}
An example of a symmetric $f$-divergence is the Bhattacharyya (BH) distance, defined as
\begin{align} \label{eq:BHdistance}
  D_{BH}(f_k, f_\ell) = - \log \int \sqrt{ {f_k(\mb{x})}{f_\ell (\mb{x})}} d\mb{x},
\end{align}
and can be a better choice if the absolute difference, and not the ratio, differentiates the two pdfs.
The appropriate divergence for a specific task can be chosen based on identified properties, e.g.\ for clustering \cite{Mollersen2016DataIndependent}, or a new dissimilarity function can be proposed \cite{Mollersen2015Divergencebased}. 

This article aims to identify properties for bag classification, and we make the following contributions:
\begin{itemize}
  \item Presenting the hierarchical model for general, non-standard MI assumptions (Section~\ref{sec:Hierarchical}).
  \item Introduction of bag-to-class dissimilarity measure (Section~\ref{sec:Bagtoclass}).
  \item Identification of two properties for bag-to-class divergence (Section~\ref{sec:Properties}).
  \item A new bag-to-class dissimilarity measure for sparse training data (Section~\ref{sec:Classconditional}).
\end{itemize}
In Section~\ref{sec:Data}, the KL information and the new dissimilarity measure is applied to data sets and the results are reported.  
Bags defined in the probability distribution space, in combination with bag-to-class divergence, constitutes a new framework for MI learning, which is compared to other frameworks in Section~\ref{sec:Discussion}.  

\section{Related work} \label{sec:Related} 

The feature vector set viewpoint seems to be the most common, but the probabilistic viewpoint was introduces already in 1998, then under the i.i.d.\ given class assumption \cite{Maron1998Framework}.
This assumption has been used in approaches such as estimating the expectation by the mean \cite{Xu2004Logistic}, or estimation of class distribution parameters \cite{Tax2011Bag}, but has also been criticised \cite{Zhou2009Multiinstance}. 
The hierarchical distribution was introduced for learnability theory under the standard MI assumption for instance classification \cite{Doran2016MultipleInstance}, and we expand the use for more general assumptions. 

Dissimilarities in MI learning have been categorised as instance-to-instance or bag-to-bag \cite{Amores2013Multiple,Cheplygina2016DissimilarityBased}.
The bag-to-prototype approach in \cite{Cheplygina2016DissimilarityBased} offers an in-between category, but the theoretical framework is missing. 
Bag-to-class dissimilarity has not been studied within the MI framework, but was used under the i.i.d.\ given class assumption for image classification in \cite{Boiman2008In}, where also the sparseness of training sets was addressed: if the instances are aggregated on class level, a denser representation is achieved.
Many MI algorithms use dissimilarities, e.g.\ graph distances \cite{Lee2012Bridging}, Hausdorff metrics \cite{Scott2005Generalized}, 
functions of the Euclidean distance \cite{Cheplygina2015Multiple, RuizMunoz2016Enhancing}, and distribution parameter based distances \cite{Cheplygina2015Multiple}.
The performances of dissimilarities on specific data sets have been investigated \cite{Cheplygina2015Multiple, Tax2011Bag,  Cheplygina2016DissimilarityBased, RuizMunoz2016Enhancing, Sorensen2010DissimilarityBased}, but more analytical comparisons are missing.
A large class of commonly used kernels are also distances \cite{Scholkopf2000Kernel}, and hence, many kernel-based approaches in MI learning can be viewed as dissimilarity-based approaches. 
In \cite{Wei2017Scalable}, the Fisher kernel is used as input to a support vector machine (SVM), whereas in \cite{Zhou2009Multiinstance} and \cite{Qiao2017Diversified} the kernels are an integrated part of the methods. 

The non-vectorial graph space was used in \cite{Zhou2009Multiinstance, Lee2012Bridging}.
We introduce the non-vectorial space of probability functions as an extension within the BS paradigm for bag classification through dissimilarity measures between distributions. 

The KL information was applied in \cite{Boiman2008In}, and is a much-used divergence function. 
It is closely connected to the Fisher information \cite{Kullback1951Information} used in \cite{Wei2017Scalable} and to the cross entropy used as loss function in \cite{Wang2018Revisiting}. 
We propose a conditional KL information in Section~\ref{sec:Classconditional}, which differs from the earlier proposed weighted KL information \cite{Sahu2003Fast} whose weight is a constant function of $X$.

\section{Theoretical background} \label{sec:Theoretical} 

\subsection{Hierarchical distributions} \label{sec:Hierarchical} 

A bag is the probability distribution from which the instances are sampled. 
The generative model of instances from a positive or negative bag follows a hierarchical distribution
\begin{align}\label{eq:GeneralHierarchical}
  \begin{aligned}
    X|B  & \sim P(X | B)  \,\,\, & X|B   \sim P(X | B)\\
    B  & \sim P(B|pos) \,\,\,  \,\,\, \text{ or } &   B  \sim P(B|neg),
     \end{aligned}
\end{align}
respectively. 
The common view in MI learning is that a bag consists of positive and negative instances, which corresponds to a bag being a mixture of a positive and a negative distribution. 

Consider tumour images labelled $pos$ or $neg$, with instances extracted from segments. 
Let $f(\mb{x}|\theta^+_k)$ and $f(\mb{x}|\theta^-_k)$ denote the pdfs of positive and negative segments, respectively, of image $k$. 
The pdf of bag $k$ is a mixture distribution
\begin{align*} 
  f_{k}(\mb{x}) = p_{k} f(\mb{x}|\theta_k^+) + (1-p_{k})f(\mb{x}|\theta_k^-),
\end{align*}
where $p_k = \sum_{i = 1}^{n_k} \tau_i/n_k$, where $\tau_i = 1$ if instance $i$ is positive. 
The probability of positive segments, $\pi_{k}$, depends on the image's class label, and hence $\pi_k$ is sampled from $P(\Pi_{pos})$ or $P(\Pi_{neg})$. 
The characteristics of positive and negative segments vary from image to image. 
Hence, $\theta^+_k$ and $\theta^-_k$ are realisations of random variables, with corresponding probability distributions $P(\Theta^+)$ and $P(\Theta^-)$. 
The generative model of instances from a positive (negative) bag is 
\begin{align} \label{eq:Hierarchical}
  \begin{split}
     X|\mathcal{T}, \Theta^+,\Theta^- & \sim \begin{cases}
      P(X|\tau = 1) = P(X|\Theta^+)\\
      P(X|\tau = 0) = P(X|\Theta^-)
    \end{cases} \\
    \mathcal{T}|\Pi_{pos(neg)} & \sim \begin{cases}
      P(\tau = 1) = \Pi_{pos(neg)}\\
      P(\tau = 0)  = 1-\Pi_{pos(neg)}
    \end{cases} \\
    \Pi_{pos(neg)} & \sim P(\Pi_{pos(neg)}), \Theta^+ \sim P(\Theta^+), \Theta^- \sim P(\Theta^-).
  \end{split}
\end{align} 
The corresponding sampling procedure from positive (negative) bag, $k$, is \\
Step 1: Draw $\pi_{k}$ from $P(\Pi_{pos(neg)})$, $\theta^+_k$ from $P(\Theta^+)$, and $\theta^-_k$ from $P(\Theta^-)$. These three parameters define the bag. \\
Step 2: For $i = 1, \ldots, n_k$, draw $\tau_i$ from $P(\mathcal{T}|\pi_{k})$, draw $\mb{x}_i$ from $P(X|\theta_k^+)$ if $\tau_i = 1$, and from $P(X|\theta_k^-)$ otherwise.

By imposing restrictions, assumptions can be accurately described, e.g.\ the standard MI assumption: 
at least one positive instance in a positive bag: $P(p_k \geq 1/n_k) = 1$; 
no positive instances in a negative bag: $P(\Pi_{neg} = 0) = 1$; 
the positive and negative instance spaces are disjoint. 

Eq.~\ref{eq:Hierarchical} is the generative model of MI problems, assuming that the instances have unknown class labels and that the distributions are parametric. 
The parameters $\pi_k$, $\theta_k^+$ and $\theta_k^-$ are i.i.d.\ samples from their respective distributions, but are not observed and are hard to estimate, due to the very nature of MI learning: The instances are not labelled.  
Instead, $P(X|B)$ can be estimated from the observed instances, and a divergence function can serve as classifier. 

\subsection{Bag-to-class dissimilarity} \label{sec:Bagtoclass}

The training set in MI learning is the instances, since the bag distributions are unknown.
Under the assumption that the instances from each bag are i.i.d.\ samples, the KL information has a special role in model selection, both from the frequentist and the Bayesian perspective.
Let $f_{bag}(\mb{x})$ be the sample distribution (unlabelled bag), and let $f_k(\mb{x})$ and $f_\ell(\mb{x})$ be two models (labelled bags). 
Then the expectation over $f_{bag}(\mb{x})$ of the log ratio of the two models, $E \{ \log ( f_k(\mb{x})/f_\ell(\mb{x})) \} $, is equal to $D_{KL}(f_{bag}, f_\ell)- D_{KL}(f_{bag}, f_k)$.
In other words, the log ratio test reveals the model closest to the sampling distribution in terms of KL information \cite{Eguchi2006Interpreting}.
From the Bayesian viewpoint, the Akaike Information Criterion (AIC) reveals the model closest to the data in terms of KL information, and is asymptotically equivalent to Bayes factor under certain assumptions \cite{Kass1995Bayes}.

The i.i.d.\ assumption is not inherent to the probability distribution viewpoint, but the asymptotic results for the KL information rely on it. 
In many applications, such as image analysis with sliding windows, the instances are best represented as dependent samples, but the dependencies are hard to estimate, and the independence assumption is often the best approximation. 
Doran and Ray \cite{Doran2016MultipleInstance} showed that the independence assumption is an approximation of dependent instances, but comes with the cost of slower convergence. 

If the bag sampling is sparse, the dissimilarity between $f_{bag}(\mb{x})$ and the labelled bags becomes somewhat arbitrary w.r.t.\ the true label of $f_{bag}(\mb{x})$.
The risk is high for ratio-based divergences such as the KL information, since $f_k(\mb{x})/f_\ell(\mb{x}) = \infty$ for $\{\mb{x}: f_\ell(\mb{x}) = 0, f_k(\mb{x}) > 0\}$. 
The bag-to-bag KL information is asymptotically the best choice of divergence function, but this is not the case for sparse training sets. 
Bag-to-class dissimilarity makes up for some of the sparseness by aggregation of instances.
Consider an image segment of colour {\it deep green}, which appears in {\it sea} images, but not in {\it sky} images, and a segment of colour {\it white}, which appears in both classes (waves and clouds). 
If the combination {\it deep green} and {\it white} does not appear in the training set, then a bag-to-bag KL information will result in infinite dissimilarity for all bags, regardless of class, but the bag-to-class KL information will be finite for the {\it sea} class. 

Let $P(X|C) = \int_{\mathcal{B}} P(X|B) dP_\mathcal{B} (B|C)$ be the probability distribution of a random vector from the bags of class $C$. 
Let $D(P(X|B),P(X|pos))$ and $D(P(X|B),P(X|neg))$ be the divergences between the unlabelled bag and each of the classes. 
Choice of divergence is not obvious, since $P(X|B)$ is different from both $P(X|pos)$ and $P(X|neg)$, but can be done by identification of properties.

\section{Properties for bag-level classification} \label{sec:Divergence}

\subsection{Properties for bag-to-class divergences} \label{sec:Properties} 

We here propose two properties for bag-to-class divergences regarding infinite bag-to-class ratio and zero instance probability.
Let  $P_{bag} = P(X|B)$, $P_{pos} = P(X|pos)$ and $P_{neg} = P(X|neg)$. 
Denote the divergence between an unlabelled bag and the reference distribution, $P_{ref}$, by $D(P_{bag}, P_{ref})$.

\begin{figure}[!h]
  \centering
    \includegraphics[width = 1\textwidth]{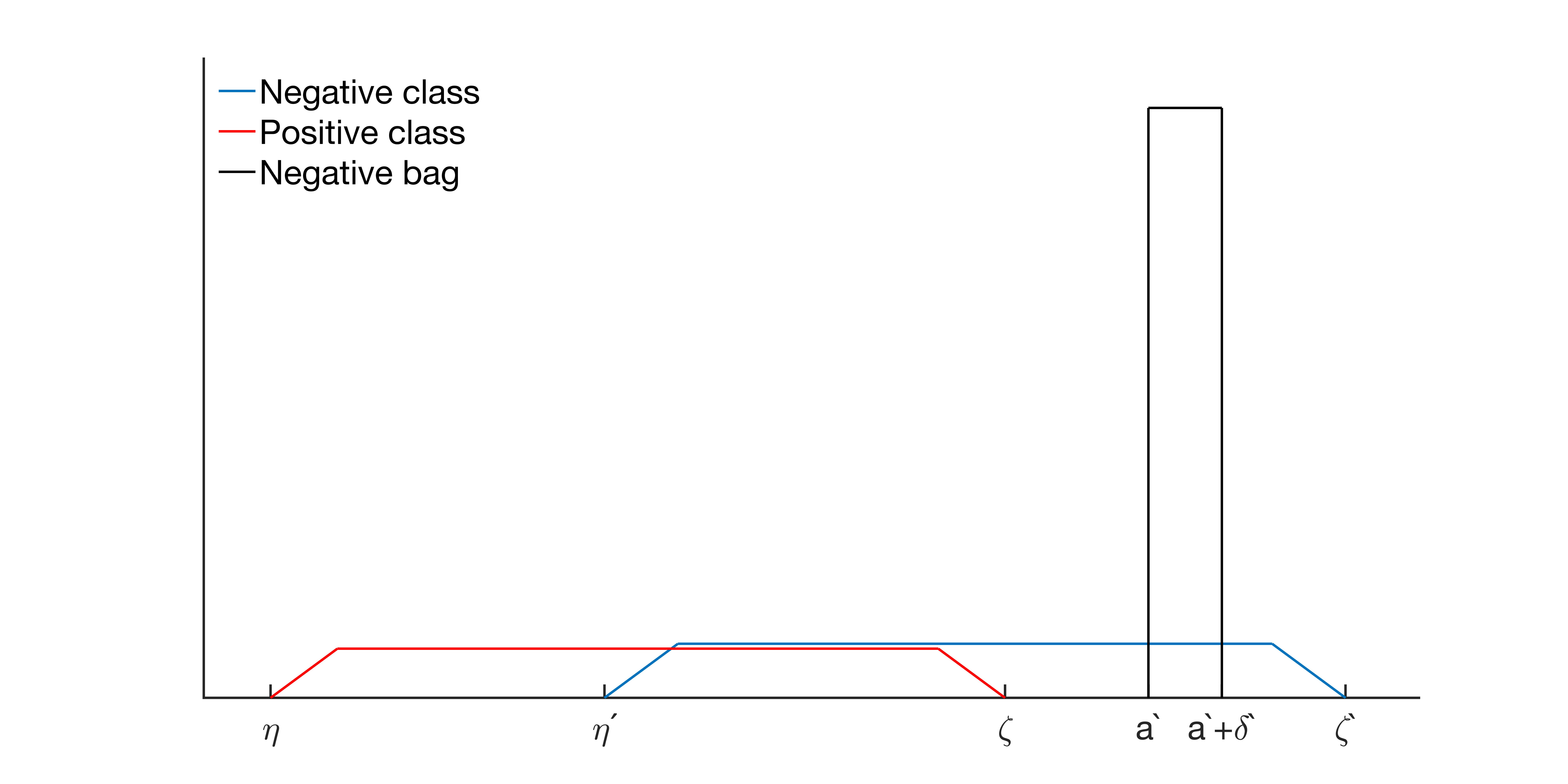}
        \caption{The pdf of a bag with uniform distribution and the pdfs of the two classes.}
        \label{fig:Uniform}
\end{figure}

As a motivating example, consider the following: A positive bag, $P_a$,  is a continuous uniform distribution $\mathcal{U}(a, a + \delta)$, sampled according to $P(A) = \mathcal{U}(\eta, \zeta - \delta)$. 
A negative bag, $P_{a'}$, is $\, \mathcal{U}(a', a'+\delta')$ sampled according to $P(A') = \mathcal{U}(\eta', \zeta'-\delta')$,  and let $ \eta' < \zeta$ so that there is an overlap between the two classes. 
For both positive and negative bags, we have that $P_{pos}/P_{bag} = \infty$ for a subspace of $\mathcal{X}$ and  $P_{neg}/P_{bag} = \infty$ for a different subspace of $\mathcal{X}$, merely reflecting that the variability in instances within a class is larger than within a bag, as illustrated in Fig.~\ref{fig:Uniform}.
If $P_{bag}$ is a sample from the negative class, and $P_{bag}/P_{pos}= \infty$ for some subspace of $\mathcal{X}$ it can easily be classified. 
From the above analysis, large bag-to-class ratio should be reflected in large divergence, whereas large class-to-bag ratio should not. 

{\bf Property 1:} For the subspace of $\mathcal{X}$ where the bag-to-class ratio is larger than some $M$, the contribution to the total divergence, $D_{\mathcal{X}_M}$, approaches the maximum contribution as $M \rightarrow \infty$. For the subspace of $\mathcal{X}$ where the class-to-bag ratio is larger than $M$, the contribution to the total divergence, $D_{\mathcal{X}_M^*}$, {\it does not} approach the maximum contribution as $M \rightarrow \infty$:   
\begin{align*}
\begin{split}
  \mathcal{X}_M: & P_{bag}/P_{ref} > M,  \, \, \, \, 
  \mathcal{X}_M^*:  P_{ref}/P_{bag} > M \\
  M  \rightarrow \infty :& 
   \begin{cases} 
    D_{\mathcal{X}_M}(P_{bag},P_{ref}) \rightarrow \max(D_{\mathcal{X}_M}(P_{bag},P_{ref})) \\
    D_{\mathcal{X}_M^*}(P_{bag},P_{ref})\centernot  \rightarrow \max(D_{\mathcal{X}_M^*}(P_{bag},P_{ref})). 
  \end{cases}
  \end{split}
\end{align*}
Property 1 can not be fulfilled by a symmetric divergence. 

As a second motivating example, consider the same positive class as before, and the two alternative negative classes defined by; 
\begin{align*}
\begin{aligned}
P_{neg} = 
   \begin{cases}
      P(A'= \eta') =  0.5\\
      P(A'= \eta'+2\delta')  =  0.5 
    \end{cases}
    \end{aligned}
    && 
    \begin{aligned}
    P_{neg'} = 
   \begin{cases}
      P(A'= \eta') =  0.5\\
      P(A'= \eta' + 2\delta')  =  0.25 \\
      P(A'= \eta' + 3\delta')  =  0.25. 
    \end{cases}
    \end{aligned}
\end{align*}
For bag classification, the question becomes: from which class is a specific bag sampled? 
It is equally probable that a bag $P_{\eta'} = P(X|A'= \eta')$ comes from each of the two negative classes, since $P_{neg}$ and $P_{neg'}$ only differ where $P_{\eta'} = 0$, and we argue that $D(P_{\eta'},P_{neg})$ should be equal to $D(P_{\eta'},P_{neg'})$. 

{\bf Property 2:} For the subspace of $\mathcal{X}$ where $P_{bag}$ is smaller than some $\epsilon$, the contribution to the total divergence, $D_{\mathcal{X}_\epsilon}$, approaches zero as $\epsilon \rightarrow 0$: 
\begin{align*}
\begin{split}
 \mathcal{X}_\epsilon &: P_{bag} < \epsilon, \, \epsilon > 0 \\
  \epsilon \rightarrow 0 & :  D_{\mathcal{X}_\epsilon}(P_{bag},P_{ref}) \rightarrow  0.
   \end{split}
\end{align*}

KL information is the only divergence that fulfils these two properties among the non-symmetric divergences listed in \cite{Taneja2006Generalized}. 
As there is no complete list of divergences, so it is possible that other divergences that the authors are not aware of fulfil these properties. 

\subsection{A class-conditional dissimilarity for MI classification} \label{sec:Classconditional} 

In the {\it sea} and {\it sky} images example, consider an unlabelled image with a {\it pink} segment, e.g.\ a boat.
If {\it pink} is absent in the training set, then the bag-to-class KL information will be infinite for both classes.
We therefore propose the following property:

{\bf Property 3:} For the subspace of $\mathcal{X}$ where both class probabilities are smaller than some $\epsilon$, the contribution to the total divergence, $D_{\mathcal{X}_\epsilon}$, approaches zero as $\epsilon \rightarrow 0$: 
\begin{align*}
\begin{split}
 \mathcal{X}_\epsilon &: P_{ref} < \epsilon, \, P'_{ref} < \epsilon \\
  \epsilon \rightarrow 0 & :  D_{\mathcal{X}_\epsilon}(P_{bag},P_{ref}) \rightarrow  0.
   \end{split}
\end{align*}

We present a class-conditional dissimilarity that accounts for this:
\begin{align} \label{eq:ClassConditional}
  {cKL}(f_{bag},f_{pos}|f_{neg}) = \int \frac{f_{neg}(\mb{x})}{f_{pos}(\mb{x})} f_{bag}(\mb{x}) \log \frac{f_{bag}(\mb{x})}{f_{pos}(\mb{x})} d\mb{x},
\end{align}
which also fulfils Properties 1 and 2. 

\subsection{Bag-level divergence classification} \label{sec:Algorithm} 

We propose two similar methods based on either the ratio of bag-to-class divergences, $rD\big(f_{bag}, f_{pos} ,f_{neg} \big)  = D\big(f_{bag}(\mb{x}), f_{pos}(\mb{x}))\big) / D\big(f_{bag}(\mb{x}),f_{neg}(\mb{x})\big )$, or the class-conditional dissimilarity in Eq.~\ref{eq:ClassConditional}. 
We propose using the KL information (Eq.~\ref{eq:KLinformation}) or the Bhattacharyya distance (Eq.~\ref{eq:BHdistance}), but any divergence function can be applied.  

Given a training set $\{(\mathbb{X}_1, y_1), \ldots, (\mathbb{X}_k, y_k), \ldots, (\mathbb{X}_K,y_K)\}$ and a set, $\mathbb{X}_{bag}$, of instances drawn from an unknown distribution, $f_{bag}(\mb{x})$, with unknown class label $y_{bag}$, and let $\mathbb{X}_{neg (pos)}$ denote the set of all $\mb{x}_{ik} \in \big (\mathbb{X}_k, y_k = neg (pos)\big )$. The bag-level divergence classification follows the steps:
\begin{align} \label{eq:Algorithm}
1. & \text{ Estimate pdfs: Fit }\hat{f}_{neg}(\mb{x}) \text{ to } \mathbb{X}_{neg} \text{, } \hat{f}_{pos}(\mb{x}) \text{ to } \mathbb{X}_{pos}, \text{ and }  \hat{f}_{bag}(\mb{x}) \text{ to } \mathbb{X}_{bag}. \nonumber\\
2. & \text{ Calculate divergences: } D\big(\hat{f}_{bag}(\mb{x}), \hat{f}_{neg}(\mb{x}))\big)  \text{ and } D\big(\hat{f}_{bag}(\mb{x}),\hat{f}_{pos}(\mb{x})\big ), \nonumber \\ 
& \text{ or }  cKL\big (\hat{f}_{bag}(\mb{x}), \hat{f}_{pos}(\mb{x})| \hat{f}_{neg}(\mb{x})\big ) \text{ by integral approximation. } \nonumber  \\
3. & \text{ Classify according to: } \\
 & y_{bag} = 
   \begin{cases}
      pos \text{ if } rD\big(\hat{f}_{bag}, \hat{f}_{pos} ,\hat{f}_{neg} \big) < t \nonumber \\
      neg \text{ otherwise. }
    \end{cases}\\
 & \text{ or } \nonumber \\
    & y_{bag} = 
   \begin{cases}
      pos \text{ if } cKL\big (\hat{f}_{bag}, \hat{f}_{pos}| \hat{f}_{neg}\big ) < t \nonumber \\
      neg \text{ otherwise. } 
    \end{cases}
\end{align}

Common methods for pdf estimation are Gaussian mixture models (GMMs) and kernel density estimation (KDE).
The integrals in step 2 are commonly approximated by importance sampling and Riemann sums. In rare cases, e.g.\ when the distributions are Gaussian, the divergences can be calculated directly. 
The threshold $t$ can be pre-defined based on, e.g.\ misclassification penalty and prior class probabilities, or estimated from the training set by leave-one-out cross-validation. 
When the feature dimension is high and the number of instances in each bag is low, pdf estimation becomes arbitrary.
A solution is to estimate separate pdfs for each dimension, calculate the corresponding divergences $D_1, \ldots, D_{Dim}$, and treat them as inputs into a classifier replacing step 3. 
Code available at https://github.com/kajsam/Bag-to-class-divergence.

\section{Experiments} \label{sec:Data} 

\subsection{Simulated data} \label{sec:Sim} 

The following study exemplifies the difference between BH distance ratio, $rBH$, KL information ratio, ${rKL}$, and $cKL$ as classifiers for sparse training data. 
The minimum dissimilarity bag-to-bag classifiers are also implemented, based on KL information and BH distance.
The number of instances from each bag is $50$, the number of bags in the training set is varied from $1$ to $25$ from each class, and the number of bags in the test set is $100$. 
Each bag and its instances are sampled as described in Eq.~\ref{eq:Hierarchical}, and the area under the receiver operating characteristic (ROC) curve (AUC) serves as performance measure.
For simplicity, we use Gaussian distributions in one dimension for {\it Sim 1}-{\it Sim 4}:
\begin{align*}
\begin{aligned}
  X^- & \sim \mathcal{N} (\mu^-, \sigma^{2-}) \\
    \mu^- &\sim \mathcal{N} (0,10)\\
    \sigma^{2-} & = |\zeta^-|, \, \zeta^- \sim \mathcal{N} (1,1) \\
    \Pi^- & = \pi^- 
   \end{aligned}
   &&
   \begin{aligned}
   X^+ & \sim \mathcal{N} (\mu^+, \sigma^{2+}) \\
      \mu^+ & \sim \mathcal{N} (\nu^+,10) \\
     \sigma^{2+} & = |\zeta^+|, \, \zeta^+ \sim \mathcal{N} (\eta^+,1)\\
   \Pi^+ & =  0.10.
   \end{aligned}
\end{align*}

\noindent {\it Sim 1:}  $\nu^+ = 15, \,  \eta^+ = 1, \, \pi^- = 0$:  
No positive instances in negative bags. \\
{\it Sim 2:} $\nu^+ = 15, \, \eta^+ = 1, \, \pi^- =  0.01$: 
Positive instances in negative bags.\\
{\it Sim 3:} $\nu^+ = 0, \, \eta^+ = 100, \, \pi^- = 0$: 
Positive and negative instances have the same expectation of the mean, but unequal variance. \\
{\it Sim 4:} $P(\nu^+= 15) = P(\nu^+ = -15) =  0.5, \,  \eta^+ = 1, \, \pi^- =  0.01$: 
Positive instances are sampled  from two distributions with unequal mean expectation.

\begin{figure}[!h]
  \centering
  \subfloat[]{
    \includegraphics[width =  1\textwidth]{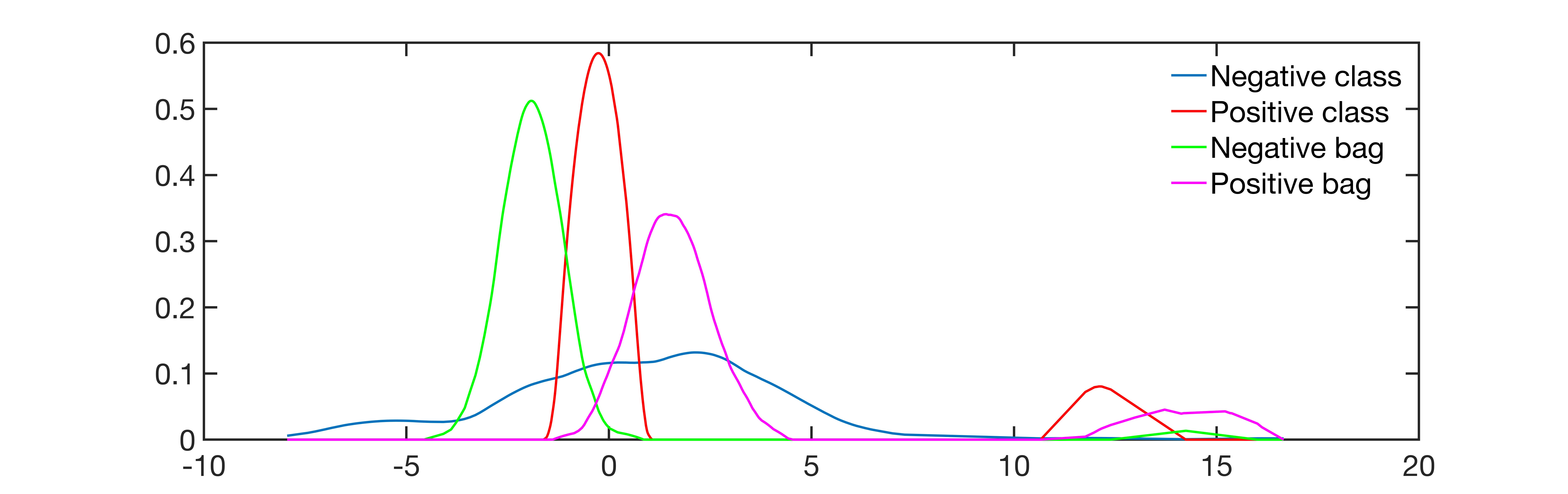}} \\
  \subfloat[]{
    \includegraphics[width =  1\textwidth]{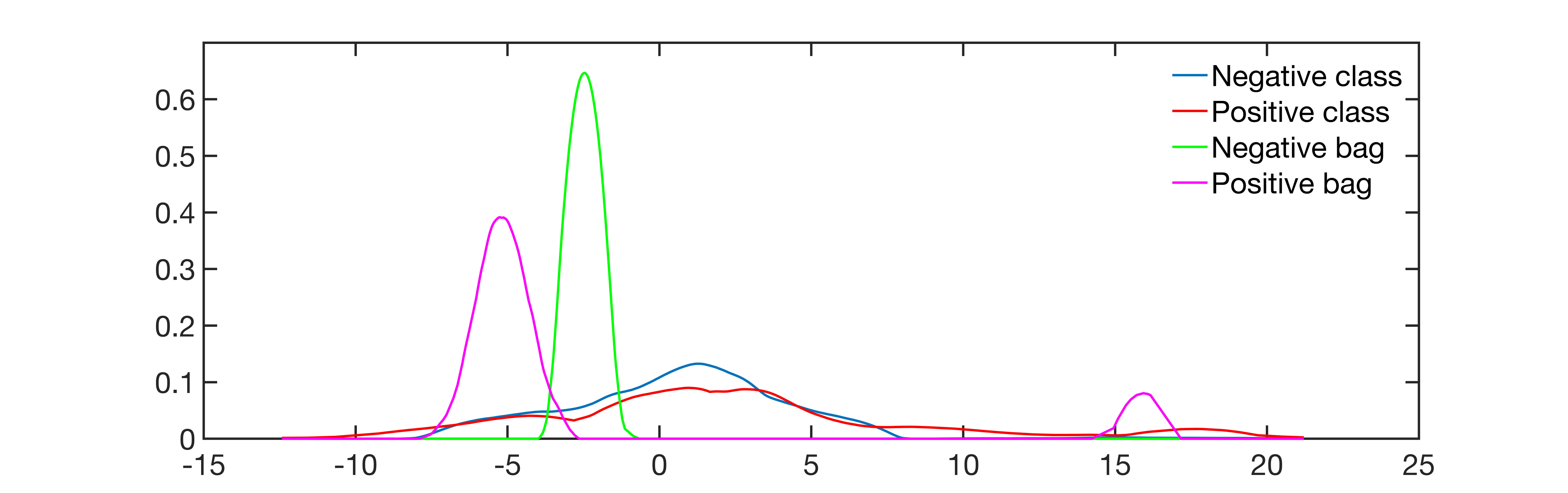}} 
    \caption{(a) One positive bag in the training set give small variance for the class pdf. (b) Ten positive bags in the training set, and the variance has increased.}
    \label{fig:SparseTraining}
\end{figure}

We add {\it Sim 5} and {\it Sim 6} for the discussion on instance labels in Section~\ref{sec:Discussion}, as follows:
{\it Sim 5} is an uncertain object classification, where the positive bags are lognormal densities with $\mu = \log(10)$ and $\sigma^2 =  0.04$, and negative bags are Gaussian mixtures densities with $\mu_1 = 9.5$, $\mu_2 = 13.5$, $\sigma^2 = 2.5$, and $\pi_1 =  0.9$. 
These two densities are nearly identical, see \cite[p.\ 15]{McLachlan2000Finite}.
In {\it Sim 6}, the parameters of {\it Sim 5} are i.i.d.\ observations from Gaussian distributions, each with $\sigma^2 = 1$ for the Gaussian mixture, and $\sigma^2 =  0.04$ for the lognormal distribution.  
Figure~\ref{fig:SparseTraining} shows the estimated class densities and two estimated bag densities for {\it Sim 2} with $10$ negative bags in the training set. 

We use the following details for the algorithm in (\ref{eq:Algorithm}): KDE fitting: Epanechnikov kernel with estimated bandwidth varying with the number of observations. Integrals: Importance sampling. 
Classifier: $t$ is varied to give the full range of sensitivities and specificities necessary to calculate AUC. 

Table~\ref{tab:Simulations1} shows the mean AUCs for $50$ repetitions.
\begin{table}[h!]
  \centering
  \caption{AUC$\cdot100$ for simulated data.}
  \label{tab:Simulations1}
  \begin{tabular}{|c|c||c|c|c||c|c|c||c|c|c|}
  \hline
   & Bags & \multicolumn{3}{c||}{$neg$: 5} & \multicolumn{3}{c||}{$neg$: 10} & \multicolumn{3}{c|}{$neg$: 25} \\
\hline
    Sim: & {$pos$}: & ${rBH}$ & ${rKL}$ & ${cKL}$ & ${rBH}$ & ${rKL}$ & ${cKL}$ & ${rBH}$ & ${rKL}$ & ${cKL}$ \\
    \hline
      & 1        &  61 &  69 &  85 &  62 &  72 &  89 &  61 &  73 &  92 \\
    1& 5        &  63 &  75 &  86 &  64 &  82 &  94 &  68 &  84 &  97 \\
    & 10        &  69 &  86 &  87 &  73 &  91 &  95 &  75 &  91 &  98 \\
    \hline
        & 1      &  57 &  61 &  75 &  59 &  61 &  78 &  58 &  55 &  75 \\
    2 & 5       &  59 &  67 &  79 &  60 &  68 &  84 &  62 &  63 &  85 \\
     &       10 &  64 &  77 &  80 &  66 &  78 &  86 &  68 &  72 &  86  \\
    \hline
    & 1           &  51 &  55 &  71 &  52 &  58 &  73 &  50 &  57 &  74 \\
    3 & 5        &  53 &  61 &  76 &  53 &  66 &  81 &  52 &  65 &  83 \\
     &       10  &  58 &  73 &  78 &  58 &  76 &  84 &  57 &  76  & 87 \\
    \hline 
     &          1 &  55 &  61 &  70 &  56 &  62 &  73 &  56 &  58 &  69 \\
    4&          5&  56 &  63 &  75 &  57 &  64 &  81 &  59 &  59 &  80 \\
    & 10         &  60 &  74 &  77 &  62 &  76 &  85 &  63 &  69 &  84 \\
    \hline
    & 1         &  64 &  61 &  62 &  67 &  63 &  66 &  64 &  62 &  67 \\
    5& 5       &  73 &  69 &  63 &  74 &  70 &  67 &  75 &  71 &  72 \\
    & 10       &  74 &  70 &  62 &  75 &  73 &  69 &  76 &  74 &  72 \\
     \hline
         & 1     &  68 &  68 &  67 &  66 &  68 &  68 &  68 &  71 &  68 \\
    6&   5      &  65 &  64 &  67 &  68 &  68 &  69 &  70 &  71 &  74 \\
    & 10        &  66 &  64 &  66 &  70 &  69 &  72 &  72 &  73 &  74 \\
    \hline
  \end{tabular}
\end{table}

\subsection{Breast tissue images} 

Breast tissue images (see Fig.~\ref{fig:Breast}) with corresponding feature vectors are used as example. 
Following the procedure in \cite{Kandemir2014Empowering}, the principal components are used for dimension reduction, and $4$-fold cross-validation is used so that $\hat{f}_{neg}(x)$ and $\hat{f}_{pos}(x)$ are fitted only to the instances in the training folds. 
For pdf estimation, GMMs are fitted to the first principal component, using an EM-algorithm, with number of components chosen by minimum AIC.
In addition, KDE as in Section~\ref{sec:Sim}, and KDE with Gaussian kernel and optimal bandwidth  \cite{Sheather1991Reliable} is used. 

\begin{table}[h!]
  \centering
  \caption{AUC$\cdot 100$ for breast tissue images.}
  \label{tab:Breast}
  \begin{tabular}{|c|c|c|c|}
 \hline
       & KDE (Epan.) & KDE (Gauss.) & GMMs \\
     \hline
     ${cKL}$ &  90 & 92 & 94 \\
     \hline
          ${rKL}$ &  82 & 92 & 96 \\
         \hline
  \end{tabular}
\end{table}

\subsection{Benchmark data} \label{sec:Benchmark} 

We here present the results for 7 benchmark datasets\footnote{https://figshare.com/articles/MIProblems\_A\_repository\_of\_multiple\_instance\_learning\_datasets/6633983} together with the results of five other methods as reported in the cited publications. 
The datasets have relatively few instances per bag compared to the dimensionality. 
For detailed descriptions and references, see \cite{Cheplygina2015Multiple}. 
We use the following details for the algorithm in (\ref{eq:Algorithm}): 
KDE fitting: Gaussian kernel with optimal bandwidth. Integrals: Importance sampling. Classifier:  Support vector machine (SVM) with linear kernel. \\
\texttt{
for d = 1: Dim \\
\indent Fit $f_{imp,d}(x)$ to $\mathbb{X}_{bag,d}$ and sample $\mb{z} = [z_{1}, \ldots, z_{n_{imp}}]$ from $f_{imp,d}(x)$. \\
\indent 1. Fit $\hat{f}_{neg,d}(x)$, $\hat{f}_{pos,d}(x)$, $\hat{f}_{bag,d}(x)$ using KDE. \\
\indent 2. Approximate $r\hat{D}_d$ or $c\hat{D}_d$ using $f_{imp,d}(x)$ and $\mb{z}$.\\
end \\
3. Input $r\hat{\mb{D}} = [r\hat{D}_1, \ldots, r\hat{D}_{Dim}]$  or $c\hat{\mb{D}} = [c\hat{D}_1, \ldots, c\hat{D}_{Dim}]$ to SVM. 
}

10 times 10-fold cross-validation is used, except for the {\it 2000-Image} dataset where 5 times 2 fold cross-validation is used as in \cite{Wei2017Scalable} and \cite{Zhou2009Multiinstance}.
In \cite{Cheplygina2016DissimilarityBased}, one 10-fold cross-validation was performed, and the standard error was reported. In \cite{Wang2018Revisiting}, 5 times 10-fold cross-validation was performed. In \cite{Qiao2017Diversified}, several parameters are optimised for each data set, which prevents a fair comparison, and there was no reported deviation/error. 
The accuracies and the standard deviations are presented in Table~\ref{tab:Benchmark1} and Table~\ref{tab:Benchmark2}, where the highest accuracies for each data set and those within one standard deviation are marked in bold.  
\begin{table}[h!]
  \centering
  \caption{Accuracy and standard deviation/error for benchmark data sets.}
  \label{tab:Benchmark1}
  \begin{tabular}{|c|c|c|c|c|c|c|}
    \hline
   & \small Musk1 & \small Musk2 & {\small Fox} & {\small Tiger} & {\small Elephant} \\
 \hline
        \footnotesize{MI-Net(DS)\cite{Wang2018Revisiting}} & \footnotesize {\bf 89.4} (9.3) & \footnotesize 87.4 (9.7) & \footnotesize 63.0 (8.0) & \footnotesize {\bf 84.5} (8.7) & \footnotesize {\bf 87.2} (7.2) \\
     \hline 
      \footnotesize{miFV$_{def}$\cite{Wei2017Scalable}} & \footnotesize {\bf 87.5} (10.6) & \footnotesize 86.1 (10.6) & \footnotesize 56.0 (9.9) & \footnotesize 78.9 (9.1) & \footnotesize 78.9 (9.1) \\
     \hline
      \footnotesize{miGraph\cite{Zhou2009Multiinstance}}  & \footnotesize {\bf 88.9} (3.3) & \footnotesize {\bf 90.3} (2.6) & \footnotesize 61.6 (2.8) &  \footnotesize {\bf 86.0} (1.6) &  \footnotesize {\bf 86.8} (0.7)\\
     \hline
       \footnotesize{$D^{RS}$\cite{Cheplygina2016DissimilarityBased}} & \footnotesize {\bf 89.3} (3.4) & \footnotesize 85.5 (4.7) & \footnotesize 64.4 (2.2) & \footnotesize 81.0 (4.6) & \footnotesize {\bf 80.4} (3.5)\\
        \hline
       \footnotesize{$DivDict$\cite{Qiao2017Diversified}} & \footnotesize 87.7 & \footnotesize 89.1 & \footnotesize 65.0 & \footnotesize 80.0 & \footnotesize 90.67 \\
     \hline
       \footnotesize{rBH} & \footnotesize 64.4 (3.1) & \footnotesize 69.2 (3.2) &  \footnotesize {\bf 71.5} (1.2) & \footnotesize  \footnotesize 70.1 (1,3) & \footnotesize {\bf 81.7} (1.7) \\
     \hline
        \footnotesize{cKL} & \footnotesize 74.0 (1.9) & \footnotesize 69.9 (2.0) & \footnotesize 65.8 (2.1) &  \footnotesize {\bf 85.0} (1.4) & \footnotesize 71.1 (3.3)\\
     \hline
    \end{tabular}
\end{table}

\begin{table}[h!]
  \centering
  \caption{Accuracy and standard deviation/error for benchmark data sets.}
  \label{tab:Benchmark2}
  \begin{tabular}{|c|c|c|c|}
    \hline
   &  {\small 2000 - Image} & {\small Alt.atheism} \\ 
 \hline
        \footnotesize{MI-Net(DS)}  & -  &  \footnotesize {\bf 86.0} (13.4)\\
    \hline
        \footnotesize{miFV$_{def}$}  & \footnotesize {\bf 87.5} (7.2)  & -   \\
     \hline
      \footnotesize{miGraph} &  \footnotesize 72.1  & \footnotesize{65.5 (4.0)} \\
     \hline
       \footnotesize{$D^{RS}$} & - & \footnotesize{44.0 (4.5)} \\
     \hline
       \footnotesize{rBH} & \footnotesize {\bf 90.0} (6.4) & \footnotesize 62.0 (2.6)\\
     \hline
        \footnotesize{cKL} &  \footnotesize 80.1 (10.5)&  \footnotesize  {\bf 85.5} (1.4)\\
     \hline
    \end{tabular}
\end{table}

\subsection{Results} \label{sec:Results} 

The general trend in Table~\ref{tab:Simulations1} is that $cKL$ gives higher AUC than $rKL$, which in turn gives higher AUC than $rBH$, in line with the divergences' properties for sparse training sets. 
The same trend can be seen with a Gaussian kernel and optimal bandwidth (numbers not reported). 
The gap between $cKL$ and $rKL$ narrows with larger training sets.
In other words, the benefit of $cKL$ increases with sparsity.
This can be explained by the $\infty/\infty$ risk of $rKL$, as seen in Figure~\ref{fig:SparseTraining}(a). 

Increasing $\pi^+$ also narrows the gap between $rKL$ and $cKL$, and eventually (at approximately $\pi^+ =  0.25$), $rKL$ outperforms $cKL$ (numbers not reported). 
{\it Sim 1} and {\it Sim 3} are less affected because the ratio $\pi^+/\pi^-$ is already $\infty$. 

\begin{figure}[!h]
  \centering
   \includegraphics[height =  0.5\textheight]{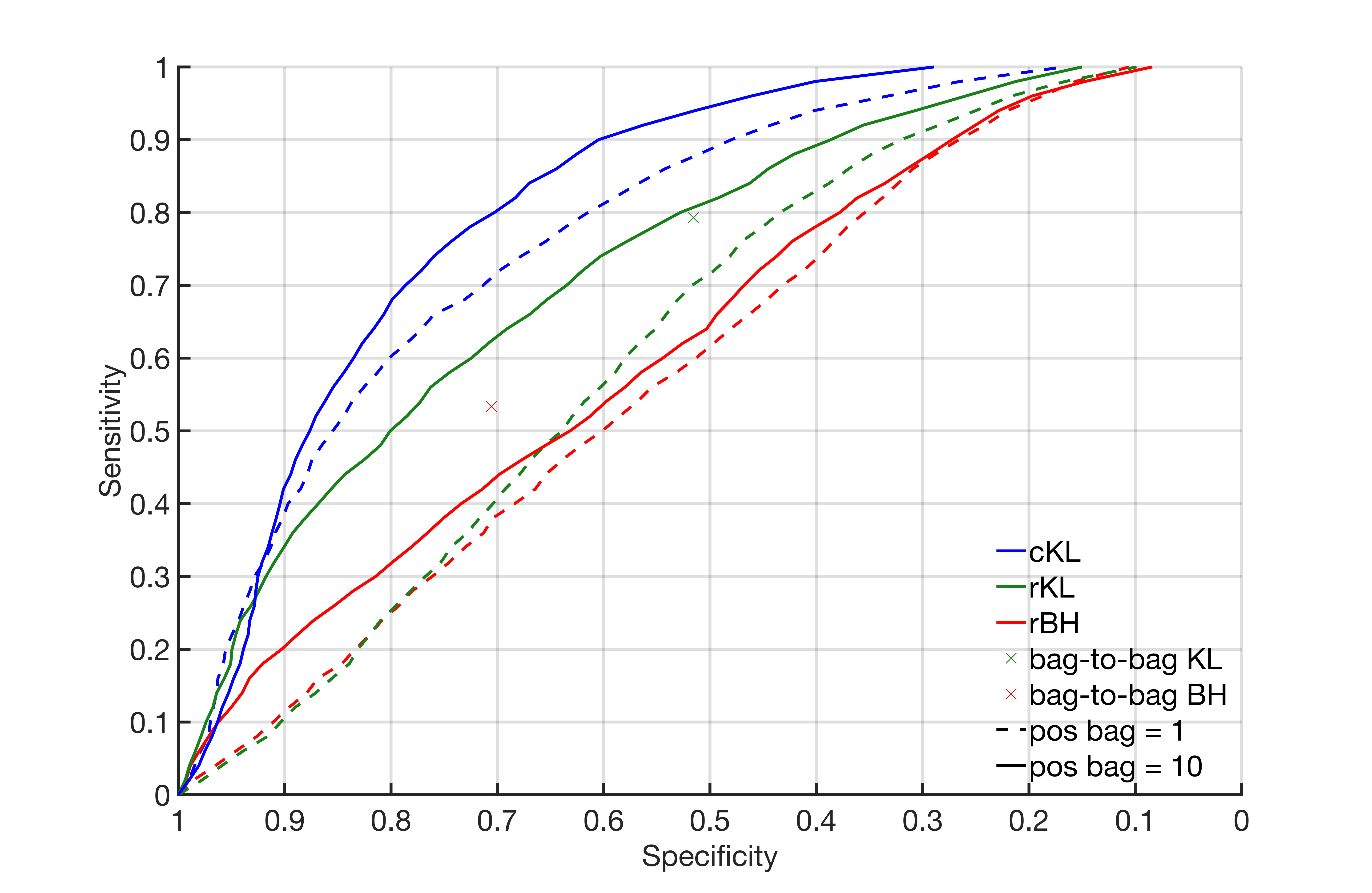}
    \caption{An example of ROC curves for $cKL$, $rKL$ and $rBH$ classifiers. The performance increases when the number of positive bags in the training set increases from $1$ (dashed line) to $10$ (solid line). The sensitivity-specificity pairs for the bag-to-bag KL and BH classifier is displayed for $100$ positive and negative bags in the training set for comparison.}
    \label{fig:ROC_SEP}
\end{figure}

The minimum bag-to-bag classifier gives a single sensitivity-specificity outcome, and the KL information outperforms the BH distance.  
Compared to the ROC curve, as illustrated in Fig.~\ref{fig:ROC_SEP}, the minimum bag-to-bag KL information classifier exceeds the bag-to-class dissimilarities only for very large training sets, typically for 500 or more, then at the expense of extensive computation time.  

{\it Sim 5} is an example in which the absolute difference, and not the ratio, differentiates the two classes, and $rBH$ has the superior performance.
When the extra hierarchy level is added in {\it Sim 6}, the performances returned to normal.

The breast tissue study shows that the simple divergence-based approach can outperform more sophisticated algorithms.
$rKL$ is more sensitive than $cKL$ to choice of density estimation method. 
$rKL$ performs better than $cKL$ with GMM, and both exceed the AUC of $0.90$ of the original algorithm. 
Table~\ref{tab:Breast} shows how the performance can vary between two common pdf estimation methods that do not assume a particular underlying distribution. 
Both KDE and GMM are sensitive to chosen parameters or parameter estimation method, bandwidth and number of components, respectively, and no method will fit all data sets. 
In general, KDE is faster, but more sensitive to bandwidth, whereas GMM is more stable. 
For bags with very few instances the benefits of GMM cannot be exploited, and KDE is preferred. 

The benchmark data study shows that the proposed method combined with a standard classifier obtain comparable results with state-of-the-art algorithms, with the exception of the Musk data sets where the number of instances per bag is low. In $Musk1$, more than half of the bags contain less than 5 instances, and in $Musk2$, one fourth of the bags contain less than 5 instances. 
Few instances per bag prevents good distribution estimations, and since the proposed method is based on bag distributions, the result is not surprising. 
The algorithms perform in the same range, although they are conceptually very different: 
{\it MI-Net} is a neural network approach, {\it miFV} is a kernel approach, {\it miGraph} is a graph approach, {\it D$^{RS}$} is a dissimilarity approach, and {\it DivDict} is a diverse dictionary approach. 

\section{Discussion} \label{sec:Discussion} 

\subsection{Point-of-view} \label{sec:Point}

The theoretical basis of the bag-to-class divergence approach relies on viewing a bag as a probability distribution, and hence fits into the branch of collective assumptions of the Foulds and Frank taxonomy \cite{Foulds2010Review}.
The probability distribution estimation can be seen as extracting bag-level information from a set $\mathbb{X}$, and hence falls into the BS paradigm of Amores \cite{Amores2013Multiple}. 
The probability distribution space is non-vectorial, different from the distance-kernel spaces in \cite{Amores2013Multiple}, and divergences are used for classification. 

In practice, the evaluation points of the importance sampling gives a mapping from the set $\mathbb{X}$ to a single vector, $\hat{f}_{bag}(\mb{z})$.
The mapping concurs with the ES paradigm, and the same applies for the graph-based methods. 
From that viewpoint, the bag-to-class divergence approach expands the distance branch of Foulds and Frank to include a bag-to-class category in addition to instance-level and bag-level distances.
However, the importance sampling is a technicality of the algorithm, and we argue that the method belongs to the BS paradigm. 
When the divergences are used as input to a classifier as in Section~\ref{sec:Benchmark}, the ES paradigm is a better description. 

Carbonneau et al.~\cite{Carbonneau2018Multiple} assume underlying instance labels, and from a probability distribution viewpoint this corresponds to posterior probabilities, which are in practice inaccessible.
In {\it Sim 1 - Sim 4}, the instance labels are inaccessible through observations without previous knowledge about the distributions.
In {\it Sim 6}, the instance label approach is not useful, due to the similarity between the two distributions: 
\begin{align} \label{eq:NoInstance}
 \begin{aligned}
    X|\Theta^+ & \sim P(X | \Theta^+)  \\
    \Theta^+ & \sim P(\Theta^+) 
  \end{aligned}
   &&
  \begin{aligned}
   X|\Theta^- & \sim P(X | \Theta^-) \\
   \Theta^-  & \sim P(\Theta^-), 
   \end{aligned}
\end{align}
where $P(X | \Theta^+)$ and $P(X | \Theta^-)$ are the lognormal and the Gaussian mixture, respectively.
Eq.~\ref{eq:Hierarchical} is just a special case of Eq.~\ref{eq:NoInstance}, where $\Theta^+$ is the random vector $\{ \Theta , \Pi_{pos}\}$.
Without knowledge about the distributions, discriminating between training sets following the generative model of Eq.~\ref{eq:Hierarchical} and Eq.~\ref{eq:NoInstance} is only possible for a limited number of problems.
Even the uncertain objects of {\it Sim 5} is difficult to discriminate from MI objects based solely on the observations in the training set. 

\subsection{Conclusions and future work} 

Although the bag-to-bag KL information has the minimum misclassification rate, the typical bag sparseness of MI training sets is an obstacle, which is partly solved by bag-to-class dissimilarities, and the proposed class-conditional KL information accounts for additional sparsity of bags. 

The bag-to-class divergence approach addresses three main challenges MI learning.
(1) Aggregation of instances according to bag label and the additional class-conditioning provides a solution for the bag sparsity problem.
(2) The bag-to-bag approach suffers from extensive computation time, solved by the bag-to-class approach. 
(3) Viewing bags as probability distributions give access to analytical tools from statistics and probability theory, and comparisons of methods can be done on a data-independent level through identification of properties. 
The properties presented here are not an extensive list, and any extra knowledge should be taken into account whenever available.

A more thorough analysis of the proposed function, $cKL$, will identify its weaknesses and strengths, and can lead to improved versions as well as alternative class-conditional dissimilarity measures and a more comprehensive tool. 

The diversity of data types, assumptions, problem characteristics, sampling sparsity, etc. is far too large for any one approach to be sufficient. 
The introduction of divergences as an alternative class of dissimilarity functions; and the bag-to-class dissimilarity as an alternative to the bag-to-bag dissimilarity, has added additional tools to the MI toolbox.

\section*{Acknowledgements}
This research did not receive any specific grant from funding agencies in the public, commercial, or not-for-profit sectors.

\section*{Bibliography}

\end{document}